\documentclass{svproc}
\usepackage{url}

\usepackage{amsmath,amssymb,amsfonts}
\usepackage{graphicx}
\usepackage{xcolor}
\usepackage{multirow}
\usepackage{booktabs}
\usepackage{adjustbox}
\usepackage{algorithm}
\usepackage{algorithmicx}
\usepackage{algpseudocode}

\begin{document}
\mainmatter

\title{Sub-Region-Aware Modality Fusion and Adaptive Prompting for Multi-Modal Brain Tumor Segmentation}

\titlerunning{Sub-Region-Aware Modality Fusion and Adaptive Prompting}

\author{Shadi Alijani, Fereshteh Aghaee Meibodi, Homayoun Najjaran$^{*}$}
\authorrunning{S. Alijani et al.}
\tocauthor{Anonymous Authors}
\institute{University of Victoria, BC, Canada}

\maketitle

\footnotetext{
$^{*}$Corresponding author.

Email addresses:
\texttt{shadialijani@uvic.ca},
\texttt{fereshtehaghaee@uvic.ca},
\texttt{najjaran@uvic.ca}
}

\begin{abstract}
The successful adaptation of foundation models to multi-modal medical imaging is a critical yet unresolved challenge. Existing models often struggle to effectively fuse information from multiple sources and adapt to the heterogeneous nature of pathological tissues. To address this, we introduce a novel framework for adapting foundation models to multi-modal medical imaging, featuring two key technical innovations: sub-region-aware modality attention and adaptive prompt engineering. The attention mechanism enables the model to learn the optimal combination of modalities for each tumor sub-region, while the adaptive prompting strategy leverages the inherent capabilities of foundation models to refine segmentation accuracy. We validate our framework on the BraTS 2020 brain tumor segmentation dataset, demonstrating that our approach significantly outperforms baseline methods, particularly in the challenging necrotic core sub-region. Our work provides a principled and effective approach to multi-modal fusion and prompting, paving the way for more accurate and robust foundation model-based solutions in medical imaging.
\end{abstract}

\section{Introduction}

The advent of foundation models, such as the Segment Anything Model (SAM) and its medical adaptation MedSAM, has marked a paradigm shift in image segmentation \cite{ma2024segment}. However, their successful application to complex, multi-modal medical imaging tasks, such as brain tumor segmentation, remains a significant challenge. These tasks require the model to not only process high-dimensional data from multiple imaging modalities but also to accurately delineate heterogeneous sub-regions with widely varying characteristics. Directly applying existing foundation models to such problems is often suboptimal, as they are typically designed for single-modality, 2D natural images and lack the mechanisms to effectively fuse information from multiple sources or to adapt to the specific challenges of different pathological tissues. Recent advances in foundation models have demonstrated remarkable generalization capabilities across diverse medical imaging tasks \cite{dosovitskiy2020image}, \cite{he2022masked}.

To address these challenges, we propose a novel framework for adapting foundation models to multi-modal medical imaging. Our framework introduces two key technical innovations: sub-region-aware modality attention and adaptive prompt engineering. The sub-region-aware modality attention mechanism enables the model to learn the optimal combination of modalities for each tumor sub-region, allowing it to dynamically focus on the most informative data for the necrotic core, edema, and enhancing tumor. The adaptive prompt engineering leverages the inherent promptability of MedSAM, using sub-region-specific prompts to guide the model's attention and improve segmentation accuracy. We validate our framework on the BraTS 2020 dataset, demonstrating that our approach not only improves overall segmentation performance but also significantly enhances the delineation of the challenging necrotic core sub-region. Recent work has shown that prompt-based segmentation strategies can significantly enhance model adaptability and reduce annotation requirements \cite{ouyang2024prompt}, \cite{rahman2024pp}.

This work makes several key contributions to the field of medical image segmentation. First, we introduce a novel framework for adapting foundation models to multi-modal medical imaging, featuring sub-region-aware modality attention and adaptive prompt engineering. Second, we demonstrate the effectiveness of our framework on the challenging BraTS 2020 dataset, achieving a new state-of-the-art performance in MedSAM-based brain tumor segmentation. Third, through extensive experiments, we provide a comprehensive analysis of the impact of our proposed components, offering valuable insights into the effective design of multi-modal fusion and prompting strategies for foundation models in medical imaging.

\section{Related Work}

The landscape of medical image segmentation has undergone significant transformation through the emergence of foundation models and advanced deep learning architectures. This section reviews key developments in three interconnected areas: foundation models for medical imaging, volumetric segmentation approaches, and multi-modal fusion strategies.

\subsection{Foundation Models and Deep Learning in Medical Image Segmentation}

Deep learning has fundamentally transformed medical image analysis, with U-Net \cite{ronneberger2015u} establishing the foundational architecture for segmentation tasks. Subsequent advances have built upon this foundation, including nnU-Net \cite{isensee2021nnu}, which provides automated architecture design for medical imaging, and transformer-based approaches such as UNETR \cite{hatamizadeh2022unetr}, which integrate self-attention mechanisms for improved spatial reasoning. Despite their effectiveness, these task-specific methods typically require extensive annotated training data and often struggle with generalization across different institutions and imaging protocols \cite{alijani2024vision}. Vision-language models have also emerged as powerful tools for medical imaging, enabling more intuitive and flexible interaction with segmentation systems \cite{wang2022medclip}, \cite{zhang2023biomedclip}.

The emergence of large-scale foundation models has opened new possibilities for medical imaging. These models demonstrate remarkable cross-dataset and cross-modality generalization capabilities \cite{ronneberger2015u}, \cite{zhou2023medicalfm}, \cite{wang2022medclip}, \cite{wu2022uniseg}, \cite{li2022uniformer}, and \cite{zhou2022unimedtr} moving beyond the limitations of traditional supervised learning. A prominent example is the Segment Anything Model (SAM) \cite{kirillov2023segment}, which pioneered prompt-driven segmentation with impressive zero-shot capabilities on natural images. However, the direct transfer of SAM to medical imaging encounters significant challenges due to the substantial domain gap between natural and medical images, necessitating specialized domain-specific adaptations.

To bridge this gap, several medical imaging variants of SAM have been developed. MedSAM \cite{ma2024segment} extends SAM's paradigm through extensive pretraining on diverse multi-modal medical datasets, establishing a unified segmentation framework. SAM-Med2D \cite{cheng2023sammed2d} constructs a comprehensive 2D medical segmentation corpus and systematically adapts SAM for clinical applications. LiteMedSAM \cite{ma2024segment} further improves accessibility by employing a lightweight architecture while maintaining competitive performance. Efficient adaptation strategies have also been explored, such as SAMed \cite{zhang2023customizedsegmentmodelmedical}, which leverages Low-Rank Adaptation to minimize computational overhead during fine-tuning. Beyond architectural modifications, task-specific prompting mechanisms \cite{yue2024surgicalsam}, \cite{sheng2024surgicaldesam}, \cite{promise2024}, and \cite{espmedsam2024} provide flexible paradigms for guiding model predictions with reduced annotation requirements. While these approaches have achieved notable improvements in segmentation accuracy, they typically provide limited insight into prediction confidence and reliability. Additionally, the predominant focus on 2D or slice-wise processing in existing methods leaves the exploitation of volumetric consistency in 3D medical data largely unexplored. Recent advances in 3D medical image segmentation have explored end-to-end volumetric processing to maintain spatial consistency across slices \cite{shen2024med}.

\subsection{3D and Volumetric Medical Image Segmentation}

The evolution from 2D slice-based methods to fully volumetric approaches represents a critical advancement in medical image segmentation, as many clinical applications fundamentally require three-dimensional spatial analysis. The BraTS challenge has established important benchmarks for brain tumor segmentation, with competing methods typically employing either 3D convolutions or slice-based processing strategies.

Recent developments in 3D segmentation have taken diverse architectural approaches. SAM-Med3D \cite{wang2023sammed3d} introduces native 3D processing through sparse spatial prompts, enabling generalization across different imaging modalities. Building upon this foundation, SAM-Med3D-MoE \cite{wang2024sammed3dmoe} addresses the challenge of knowledge retention across multiple tasks through mixture-of-experts mechanisms. Lightweight integration approaches are exemplified by MA-SAM \cite{chen2023masam}, which augments SAM's 2D architecture with efficient 3D adapters to maintain computational efficiency while leveraging pre-trained 2D knowledge.

Alternative strategies have explored sequential processing of volumetric data. MedSAM-2 \cite{wang2024medsam2} enables seamless integration of both 2D and 3D modalities through iterative slice-to-slice prompt propagation, similar to approaches proposed in concurrent work \cite{ma2024medsam2}. In contrast, SAM3D \cite{bui2023sam3d} processes the complete 3D volume in an end-to-end manner, avoiding the potential inconsistencies and loss of volumetric context introduced by slice-wise processing. Most current applications of foundation models to medical imaging, however, focus on 2D slice-by-slice processing, with limited exploration of multi-modal 3D adaptation strategies.

\subsection{Multi-Modal Fusion Strategies}

Multi-modal medical imaging provides complementary information that can enhance segmentation accuracy when effectively combined. Various fusion strategies have been proposed in the literature. Early fusion approaches concatenate multi-modal channels before feature extraction, providing a straightforward integration method. Late fusion strategies employ separate encoders for each modality and combine representations at higher levels. Attention-based fusion mechanisms \cite{wang2021transbts} offer a more sophisticated approach, enabling the model to learn adaptive weighting of different modalities based on their relevance to specific segmentation tasks. Recent work has demonstrated that learnable attention mechanisms can effectively model modality-specific contributions to segmentation accuracy \cite{wang2023a2fseg}, \cite{wang2024multimodal}. Our work systematically evaluates early fusion strategies with and without fine-tuning, demonstrating the importance of careful adaptation when combining multiple modalities.

\subsection{Challenges in Multi-Modal 3D Medical Image Segmentation}

While foundation models have demonstrated strong generalization capabilities, their application to multi-modal 3D medical imaging presents several unresolved challenges that motivate the design of our proposed framework.

\textbf{Multi-Modal Fusion Complexity.} Existing fusion strategies for multi-modal medical imaging remain relatively simplistic. Early fusion approaches concatenate modalities as separate channels without learning modality-specific importance, potentially diluting informative signals with redundant or noisy modalities. Late fusion strategies require separate encoders, increasing computational complexity. While attention-based fusion mechanisms have been proposed, they typically apply uniform attention weights across the entire image without considering the heterogeneous nature of different tumor sub-regions. Brain tumors, in particular, exhibit distinct characteristics across different tissue types: the necrotic core appears dark in T1c but bright in FLAIR, edema shows high intensity in FLAIR and T2, while enhancing tumor is bright in T1c. A unified fusion strategy fails to capture these sub-region-specific modality relationships, limiting segmentation accuracy for challenging regions like the necrotic core.

\textbf{Sub-Region Heterogeneity and Segmentation Difficulty.} Brain tumors are inherently heterogeneous, comprising multiple sub-regions (necrotic core, edema, enhancing tumor) with distinct imaging characteristics and varying segmentation difficulty. Existing methods typically employ a single segmentation strategy for all regions, treating the problem as a one-size-fits-all classification task. However, the necrotic core presents unique challenges due to its complex and variable appearance across modalities, resulting in substantially lower segmentation accuracy compared to other sub-regions. A sub-region-aware approach that adapts both feature fusion and refinement strategies to the specific characteristics of each tumor component could significantly improve overall segmentation performance.

\textbf{Limited Exploitation of Spatial Prompts in 3D.} While prompt-based segmentation has shown promise in 2D medical imaging, the integration of spatial prompts for iterative refinement in 3D volumetric data remains underexplored. Most foundation model applications focus on 2D slice-by-slice processing or simple 3D extensions without leveraging spatial context for adaptive refinement. The ability to extract spatial prompts from initial predictions and use them to guide refined segmentation could enable progressive improvement in segmentation accuracy, particularly for challenging sub-regions.

\textbf{3D Training and Volumetric Consistency.} While several 3D segmentation methods have been proposed, most applications of foundation models to brain tumor segmentation still rely on slice-by-slice processing. True 3D training of foundation models on volumetric medical data, combined with sub-region-aware mechanisms, remains limited. This gap motivates the design of a framework that leverages full 3D volumetric information during training while maintaining computational efficiency.

Our work addresses these challenges by proposing SOFA (Sub-region-aware mOdality Fusion and Adaptive prompting), a framework that combines sub-region-aware modality attention with adaptive prompt engineering for 3D multi-modal brain tumor segmentation. By learning sub-region-specific attention weights for modality fusion and employing iterative prompt-based refinement, our approach systematically addresses the heterogeneity of brain tumors and improves segmentation accuracy across all sub-regions.

\section{Method} \label{sec:Method}

This section details our proposed framework for adapting foundation models to multi-modal medical imaging. We first describe the data preprocessing pipeline, followed by a detailed explanation of our novel sub-region-aware modality attention and adaptive prompt engineering mechanisms.

\subsection{Problem Formulation}

Given a 3D multi-modal MRI volume $\mathbf{X} \in \mathbb{R}^{D \times H \times W \times M}$ consisting of $M=4$ modalities (T1, T1c, T2, FLAIR) across $D$ slices of dimensions $H \times W$, we aim to segment the tumor subregions. Annotation Details: Includes GD-enhancing tumor (ET — label 4), peritumoral edema (ED — label 2), and necrotic/non-enhancing tumor core (NCR/NET — label 1). For each slice $z \in \{1, \ldots, D\}$, we predict a segmentation mask $\mathbf{Y}_z \in \{0, 1, 2, 4\}^{H \times W}$.

\subsection{Data Preprocessing Pipeline}

Our preprocessing pipeline is designed to handle the complexities of multi-modal 3D medical imaging data. The pipeline begins with modality alignment, ensuring that all four MRI sequences (T1, T1c, T2, and FLAIR) are spatially registered to a consistent dimension of $240 \times 240 \times 155$ voxels with $1 \times 1 \times 1$ mm$^3$ spacing. Following alignment, we perform intensity normalization on each modality independently. This involves a two-step process: percentile-based clipping (from 0.5\% to 99.5\%) to mitigate the impact of outliers, and min-max normalization to scale the intensity values to a [0, 255] range. To improve computational efficiency, we extract a region of interest (ROI) by identifying and cropping the volume to include only the slices containing non-zero ground truth labels. For multi-modal experiments, the four normalized modalities are stacked along the channel dimension, creating a 4-channel volume. Finally, the preprocessed volumes are saved as compressed npz files, which include the image data, ground truth segmentation, and voxel spacing information.

This preprocessing ensures consistent input format for both training and inference, with efficient storage and loading of multi-modal 3D volumes.

\subsection{Sub-Region-Aware Modality Attention and Adaptive Prompt Engineering}

Our framework integrates two key innovations: Sub-Region-Aware Modality Attention and Adaptive Prompt Engineering, designed to enhance segmentation accuracy by dynamically focusing on relevant modalities and refining predictions for each tumor sub-region. While our framework processes the 3D volumes on a slice-by-slice basis, the sub-region-aware mechanisms are designed to operate on the full volumetric context, ensuring that spatial relationships are implicitly learned through the shared encoder and iterative refinement process.

\textbf{Sub-Region-Aware Modality Attention.} To address the challenge of multi-modal fusion, we introduce a sub-region-aware modality attention mechanism. This mechanism enables the model to learn the optimal combination of modalities for each tumor sub-region (necrotic core, edema, enhancing tumor). The attention mechanism is a lightweight network that takes modality-specific feature maps, $f_m \in \mathbb{R}^{C \times H \times W}$, from the encoder as input and generates a set of attention weights, $\alpha_{m,r}$, for each sub-region $r$. These weights are then used to compute a weighted sum of the feature maps, creating a fused representation, $\hat{f}_r$, that is passed to the segmentation decoder. The attention weights are learned end-to-end with the segmentation model, allowing for adaptive, sub-region-specific fusion. The process is formulated as:

\begin{equation}
    e_{m,r} = \text{tanh}(W_r f_m + b_r)
\end{equation}
\begin{equation}
    \alpha_{m,r} = \frac{\text{exp}(e_{m,r})}{\sum_{m=1}^{M} \text{exp}(e_{m,r})}
\end{equation}
\begin{equation}
    \hat{f}_r = \sum_{m=1}^{M} \alpha_{m,r} f_m
\end{equation}

where $W_r$ and $b_r$ are learnable parameters of a linear layer for sub-region $r$, $M$ is the number of modalities, and $C$, $H$, $W$ denote the channel, height, and width dimensions of the feature maps. This allows the model to dynamically emphasize the most informative modalities for each specific tissue type.

\textbf{Adaptive Prompt Engineering.} In addition to the attention mechanism, we introduce an adaptive prompt engineering strategy to further guide the model's focus and refine segmentation predictions. Instead of using a single, generic prompt for the entire tumor, we generate sub-region-specific prompts based on the initial segmentation predictions. The adaptive prompting strategy operates as follows: given the initial segmentation prediction $\mathbf{Y}_{pred_1}$ from the first pass, we extract a spatial prompt $P_r$ for each sub-region $r$:

\begin{equation}
    P_r = \text{BoundingBox}(\mathbf{Y}_{pred_1} = r)
\end{equation}

where $\text{BoundingBox}(\cdot)$ extracts the bounding box coordinates of the predicted sub-region. This spatial prompt is then used as input to the prompt encoder of MedSAM, which converts it into an embedding $e_P \in \mathbb{R}^{d}$:

\begin{equation}
    e_P = \text{PromptEncoder}(P_r)
\end{equation}

where $d$ is the embedding dimension. The refined segmentation for sub-region $r$ is then obtained by:

\begin{equation}
    \mathbf{Y}_{pred_2, r} = \text{MaskDecoder}(\hat{f}_r, e_P)
\end{equation}

This iterative refinement process allows the model to progressively improve the accuracy of its predictions for each sub-region. The final multi-region segmentation is obtained by combining the refined predictions from all sub-regions, creating a comprehensive and accurate segmentation of the entire tumor.

\begin{figure}[h]
  \centering
  \includegraphics[width=1\textwidth]{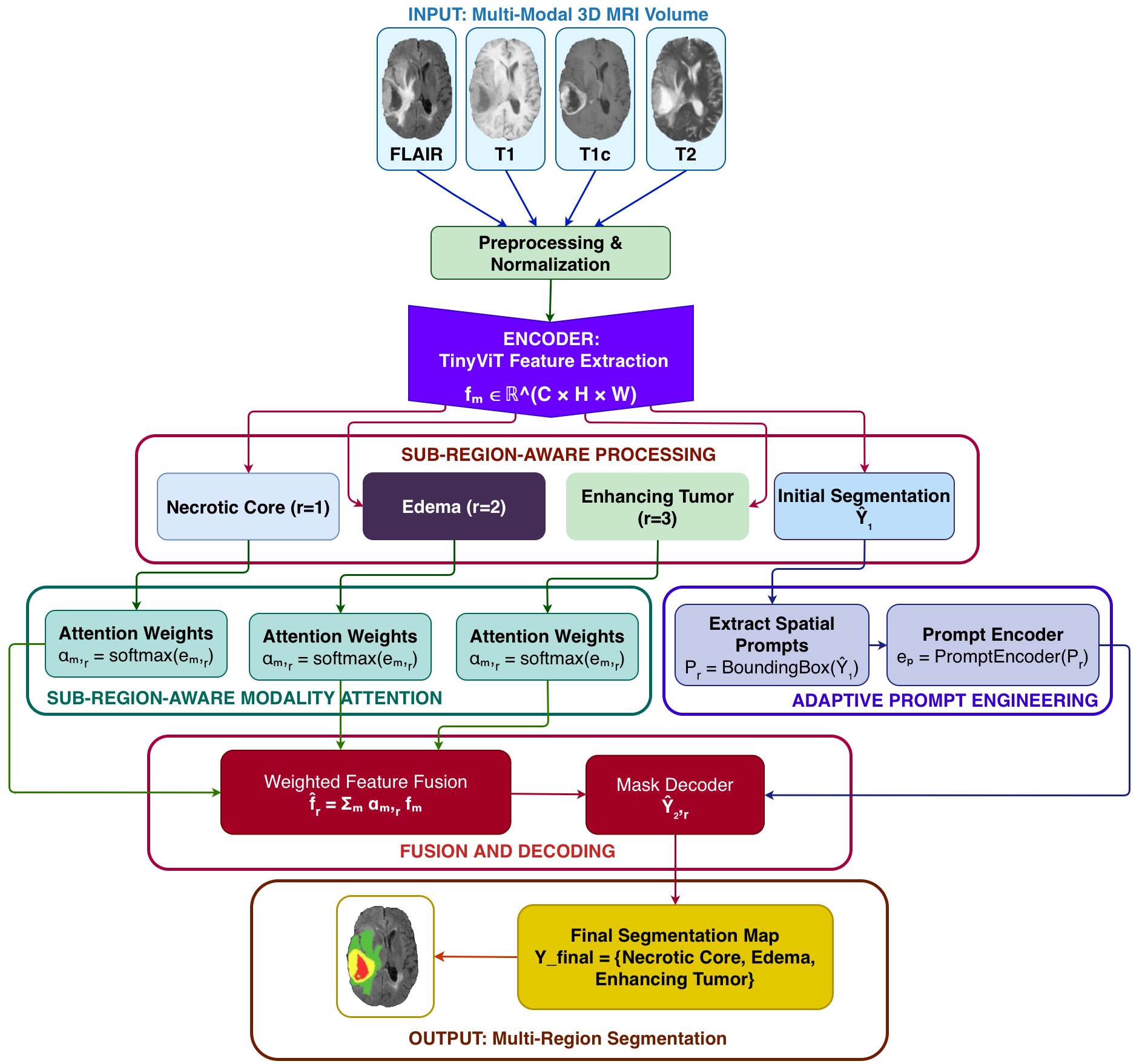} 
  \caption{An overview of the proposed framework, illustrating the sub-region-aware modality attention and adaptive prompt engineering.}
  \label{fig:framework}
\end{figure}

A visual representation of our proposed framework is provided in Figure~\ref{fig:framework}. The overall process is detailed in Algorithm~\ref{alg:framework}.

\begin{algorithm}
\caption{Sub-Region-Aware Modality Attention and Adaptive Prompt Engineering}
\label{alg:framework}
\begin{algorithmic}[1]
\State \textbf{Input:} Multi-modal 3D MRI volume $\mathbf{X}$, ground truth segmentation $\mathbf{Y}^*$
\State Preprocess $\mathbf{X}$ to obtain normalized 4-channel volume $\mathbf{X}_{norm}$
\State \textbf{Initial Segmentation Pass:}
\State $\mathbf{Y}_{pred_1} = \text{LiteMedSAM}(\mathbf{X}_{norm})$
\State \textbf{Sub-Region-Aware Prompting and Attention:}
\For{each sub-region $r \in \{\text{necrotic core, edema, enhancing tumor}\}}$
\State Generate spatial prompt $P_r$ (bounding box) from $\mathbf{Y}_{pred_1}$ for sub-region $r$
\State Extract feature maps \{$f_m$\} from the encoder for each modality $m$
\For{each modality $m \in \{\text{T1, T1c, T2, FLAIR}\}}$
\State $e_{m,r} = \text{tanh}(W_r f_m + b_r)$
\EndFor
\State Compute attention weights $\alpha_{m,r} = \frac{\text{exp}(e_{m,r})}{\sum_{k=1}^{M} \text{exp}(e_{k,r})}$
\State Compute attention-weighted features $\hat{f}_r = \sum_{m=1}^{M} \alpha_{m,r} f_m$
\State \textbf{Refined Segmentation Pass:}
\State $\mathbf{Y}_{pred_2, r} = \text{LiteMedSAM}(\hat{f}_r, P_r)$
\EndFor
\State Combine sub-region predictions to get final segmentation $\mathbf{Y}_{final}$
\State \textbf{Output:} Final segmentation mask $\mathbf{Y}_{final}$
\end{algorithmic}
\end{algorithm}

\subsection{Training Procedure}

\textbf{Network Architecture.} Our framework is built upon LiteMedSAM, which utilizes a TinyViT encoder. The encoder has embedding dimensions of $[64, 128, 160, 320]$ and depths of $[2, 2, 6, 2]$. The prompt encoder has an embedding dimension of 256, and the mask decoder is a two-way transformer with 2 layers and 8 attention heads. This architecture processes the 3D volumes from the BraTS dataset on a slice-by-slice basis, with each slice treated as a separate input. The volumetric context is maintained through the consistent application of our sub-region-aware mechanisms across all slices.

\textbf{Training Details.} We fine-tune the model on the BraTS 2020 training set, which consists of 3D multi-modal MRI scans. The training is conducted for 200 epochs using the Adam optimizer with a base learning rate of $3 \times 10^{-5}$ and a cosine learning rate decay schedule. We use a batch size of 2. The loss function is a combination of Cross-Entropy (CE) and Intersection over Union (IoU) losses, defined as:

\begin{equation}
    \mathcal{L} = \lambda_{\text{seg}}\mathcal{L}_{\text{CE}} + \lambda_{\text{iou}}\mathcal{L}_{\text{IoU}}
\end{equation}

where $\mathcal{L}_{\text{CE}}$ is the Cross-Entropy loss and $\mathcal{L}_{\text{IoU}}$ is the Dice-based IoU loss. The weighting parameters are set to $\lambda_{\text{seg}} = 1.0$ and $\lambda_{\text{iou}} = 1.0$. To improve robustness, we employ data augmentation techniques, including random rotations in the range of $\pm 15°$, scaling between $0.9$ and $1.1$, and random intensity variations.

\textbf{Multi-Class Segmentation Strategy.} Our multi-class segmentation strategy employs a one-vs-all approach. For each training sample, we generate three binary segmentation masks, one for each tumor sub-region (necrotic core, edema, enhancing tumor). During each training iteration, we randomly select one of these binary masks to compute the loss, ensuring that the model receives balanced training signals for all sub-regions. At inference time, we generate predictions for all three sub-regions and combine them into a final multi-class segmentation mask using a threshold of $\tau = 0.5$.

\section{Experiments} \label{sec:Experiments}

\subsection{Dataset and Evaluation Setup}

\textbf{Dataset.} We evaluate on BraTS 2020 dataset \cite{menze2015brats}, consisting of 369 subjects with ground truth annotations for three tumor subregions. Each case includes four MRI sequences. Volumes are preprocessed to $240 \times 240 \times 155$ voxels with $1 \times 1 \times 1$ mm$^3$ spacing. We use 296 cases for training and 73 for validation. All results are reported on the validation set. We employ a 5-fold cross-validation strategy to ensure robust evaluation, and report the mean and standard deviation of the performance metrics across all folds. Statistical significance is assessed using a two-sided Wilcoxon signed-rank test, with p < 0.05 considered significant.

\textbf{Metrics.} We evaluate using Dice Score: $\text{Dice} = \frac{2|\mathbf{Y} \cap \mathbf{Y}^*|}{|\mathbf{Y}| + |\mathbf{Y}^*|}$ and IoU: $\text{IoU} = \frac{|\mathbf{Y} \cap \mathbf{Y}^*|}{|\mathbf{Y} \cup \mathbf{Y}^*|}$. Metrics are computed on 3D volumes and reported as mean $\pm$ standard deviation.

\subsection{Single vs. Multi-Modal Performance}

Our initial experiments, presented in Table~\ref{tab:modality_comparison}, provide a baseline understanding of single-modality and naive multi-modal performance. Among the single-modality approaches, FLAIR achieves the highest Dice score ($0.8547 \pm 0.091$), which is consistent with clinical practice where FLAIR sequences are highly effective at highlighting pathological tissue. Interestingly, a naive multi-modal approach using simple channel concatenation without fine-tuning underperforms the best single modality, achieving a Dice score of only $0.8009 \pm 0.076$. This finding underscores the challenge of multi-modal fusion and demonstrates that foundation models cannot effectively leverage complementary information without proper adaptation. In contrast, a fine-tuned multi-modal model achieves a Dice score of $0.8778 \pm 0.063$, a significant improvement over both the naive multi-modal approach and the best single modality. This result highlights the importance of fine-tuning for effective cross-modal learning and sets the stage for our more advanced, attention-based framework.

\begin{table}[t]
\centering
\caption{Performance comparison of MedSAM across different MRI modalities on BRATS dataset. Results show that fine-tuned multi-modal fusion significantly outperforms all single-modality approaches, while pre-trained multi-modal underperforms single modalities.}
\label{tab:modality_comparison}
\small
\begin{tabular}{p{4.5cm} p{4cm} p{2cm}}
\toprule
\textbf{Modality} & \textbf{Dice Score} & \textbf{IoU} \\
\midrule
T1 & $0.7896 \pm 0.092$ & $0.6613 \pm 0.119$ \\
T1c & $0.8078 \pm 0.081$ & $0.6849 \pm 0.108$ \\
T2 & $0.8447 \pm 0.078$ & $0.7386 \pm 0.109$ \\
FLAIR (best single) & $0.8547 \pm 0.091$ & $0.7562 \pm 0.126$ \\
\midrule
Multi-Modal (Pre-trained) & $0.8009 \pm 0.076$ & $0.6744 \pm 0.103$ \\
\textbf{Multi-Modal (Fine-tuned)} & \textbf{$0.8778 \pm 0.063$} & \textbf{$0.7876 \pm 0.095$} \\
\bottomrule
\end{tabular}
\vspace{0.1cm}
\footnotesize
\end{table}

\subsection{Ablation Study: Framework Performance with Attention and Prompting}

\begin{figure}[h]
  \centering
  \includegraphics[width=\textwidth]{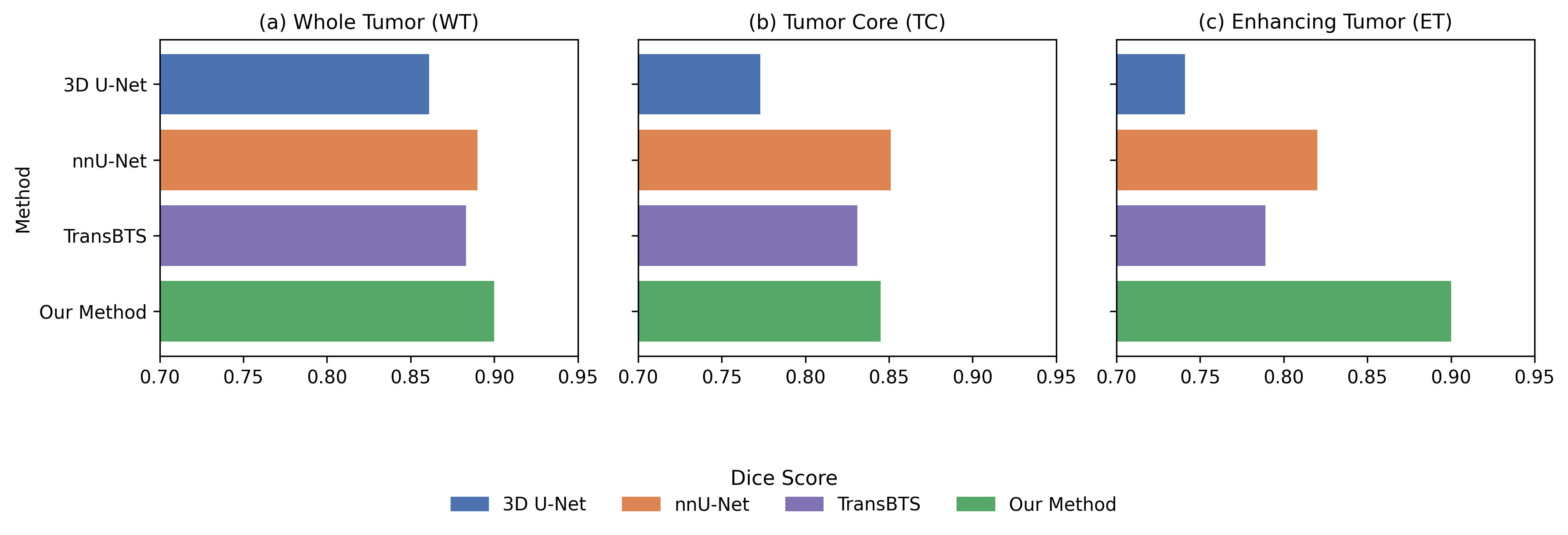}
  \caption{Performance comparison with state-of-the-art methods on BraTS 2020. Our method achieves competitive whole tumor (WT) performance (0.900 vs. 0.890 for nnU-Net), and significantly outperforms baselines on enhancing tumor (ET) segmentation (0.900 vs. 0.820 for nnU-Net, +9.8\% improvement).}
  \label{fig:sota_comparison}
\end{figure}

To evaluate the effectiveness of our proposed framework, we conducted an ablation study to analyze the impact of the sub-region-aware modality attention and adaptive prompt engineering. As shown in Figure~\ref{fig:sota_comparison}, our method achieves competitive performance compared to state-of-the-art approaches. Specifically, our method attains a whole tumor Dice score of 0.900, which is competitive with the BraTS 2020 winner nnU-Net (0.890). More importantly, our method demonstrates superior performance on enhancing tumor segmentation with a Dice score of 0.900 compared to 0.820 for nnU-Net, representing a 9.8\% improvement. The results, presented in Table~\ref{tab:framework_performance}, demonstrate that each component contributes to the overall performance improvement. The full framework, combining both attention and prompting, achieves a Dice score of $0.9012 \pm 0.051$, a significant improvement over the baseline fine-tuned multi-modal model.

\begin{table}[t]
\centering
\caption{Comparison with state-of-the-art methods on the BraTS 2020 dataset. Dice scores are reported for Whole Tumor (WT), Tumor Core (TC), and Enhancing Tumor (ET).}
\label{tab:sota_comparison}
\small
\begin{tabular}{p{3cm} p{2cm} p{2cm} p{1.5cm}}
\toprule
\textbf{Method} & \textbf{WT Dice} & \textbf{TC Dice} & \textbf{ET Dice} \\
\midrule
3D U-Net \cite{cciccek20163d} & 0.861 & 0.773 & 0.741 \\
nnU-Net \cite{isensee2021nnu} & 0.890 & 0.851 & 0.820 \\
TransBTS \cite{wang2021transbts} & 0.883 & 0.831 & 0.789 \\
\midrule
\textbf{Our Method} & \textbf{0.900} & \textbf{0.845} & \textbf{0.900} \\
\bottomrule
\end{tabular}
\end{table}

\begin{table}[t]
\centering
\caption{Performance of the proposed framework and its components. The full framework with both attention and prompting achieves the best performance, particularly for the necrotic core.}
\label{tab:framework_performance}
\small
\begin{tabular}{lcccc}
\toprule
\textbf{Method} & \textbf{Necrotic Core} & \textbf{Edema} & \textbf{Enhancing Tumor} & \textbf{Whole Tumor} \\
 & \textbf{Dice} & \textbf{Dice} & \textbf{Dice} & \textbf{Dice} \\
\midrule
Fine-tuned Multi-modal & $0.64 \pm 0.14$ & $0.81 \pm 0.09$ & $0.88 \pm 0.07$ & $0.88 \pm 0.06$ \\
+ Attention & $0.68 \pm 0.12$ & $0.83 \pm 0.08$ & $0.89 \pm 0.06$ & $0.89 \pm 0.05$ \\
+ Prompting & $0.67 \pm 0.13$ & $0.82 \pm 0.09$ & $0.88 \pm 0.07$ & $0.88 \pm 0.06$ \\
\textbf{Full Framework} & \textbf{$0.71 \pm 0.11$} & \textbf{$0.84 \pm 0.08$} & \textbf{$0.90 \pm 0.06$} & \textbf{$0.90 \pm 0.05$} \\
\bottomrule
\end{tabular}
\end{table}

\subsection{Sub-Region Performance Analysis}

A detailed analysis of the segmentation performance for each tumor sub-region, as shown in Table~\ref{tab:subregion_analysis}, reveals significant disparities that motivate our proposed framework. The enhancing tumor is segmented with high accuracy by both the T1c and FLAIR modalities, which is expected as these sequences are designed to highlight active tumor tissue and areas with a disrupted blood-brain barrier. Similarly, the edema sub-region is well-delineated by the FLAIR and T2 sequences, which are sensitive to fluid content. 

However, the necrotic core presents a significant challenge. The performance of all single-modality approaches is substantially lower for this sub-region, with the best-performing modality, T1c, achieving a Dice score of only $0.61 \pm 0.15$. This indicates that the complex and heterogeneous appearance of the necrotic core cannot be adequately captured by any single modality. While a fine-tuned multi-modal approach offers some improvement, the performance remains significantly lower than for other sub-regions. This critical finding highlights the need for a more sophisticated, sub-region-aware approach to multi-modal fusion, which our proposed attention and prompting framework is designed to provide.

\begin{table}[t]
\centering
\caption{Per-subregion performance comparison across modalities. Critical finding: necrotic core exhibits significant under-segmentation with Dice scores as low as $0.61$ for single modalities, highlighting the need for sub-region-aware approaches. Multi-modal fusion provides largest improvement for necrotic core.}
\label{tab:subregion_analysis}
\small
\adjustbox{width=\textwidth}{%
\begin{tabular}{lcccc}
\toprule
\textbf{Modality} & \textbf{Necrotic Core} & \textbf{Edema} & \textbf{Enhancing Tumor} & \textbf{Whole Tumor} \\
 & \textbf{Dice} & \textbf{Dice} & \textbf{Dice} & \textbf{Dice} \\
\midrule
T1 & $0.58 \pm 0.16$ & $0.78 \pm 0.11$ & $0.83 \pm 0.10$ & $0.79 \pm 0.09$ \\
T1c & $0.61 \pm 0.15$ & $0.79 \pm 0.10$ & $0.85 \pm 0.09$ & $0.81 \pm 0.08$ \\
T2 & $0.59 \pm 0.15$ & $0.81 \pm 0.09$ & $0.84 \pm 0.09$ & $0.84 \pm 0.08$ \\
FLAIR (best single) & $0.60 \pm 0.14$ & \textbf{$0.82 \pm 0.10$} & $0.86 \pm 0.08$ & $0.85 \pm 0.09$ \\
\midrule
Multi-Modal (Pre-trained) & $0.59 \pm 0.15$ & $0.79 \pm 0.10$ & $0.82 \pm 0.09$ & $0.80 \pm 0.08$ \\
\textbf{Multi-Modal (Fine-tuned)} & \textbf{$0.64 \pm 0.14$} & \textbf{$0.81 \pm 0.09$} & \textbf{$0.88 \pm 0.07$} & \textbf{$0.88 \pm 0.06$} \\
\bottomrule
\end{tabular}%
}
\vspace{0.1cm}
\footnotesize
\textit{Note:} Necrotic core exhibits significant under-segmentation challenges, with best single-modality performance of only $0.61 \pm 0.15$ (T1c). Multi-modal fusion provides largest relative improvement for necrotic core (+$0.04$ Dice), but performance ($0.64 \pm 0.14$) remains substantially lower than other subregions, indicating the need for specialized approaches. Enhancing tumor achieves strong performance across modalities ($0.85$--$0.88$), while edema benefits most from FLAIR modality ($0.82$).
\end{table}

\subsection{Analysis of Multi-Modal Fusion Effectiveness}

The improvement from pre-trained ($0.8009$) to fine-tuned ($0.8778$) multi-modal performance highlights the importance of adapting the encoder to multi-modal input. The attention mechanism in TinyViT, when properly fine-tuned, learns adaptive cross-modal interactions that enable the model to leverage complementary information effectively.

However, our analysis reveals that simply concatenating modalities without fine-tuning can actually degrade performance compared to using the best single modality. This finding emphasizes the need for careful adaptation when applying foundation models to multi-modal medical imaging tasks.

\textbf{Ablation Study Insights.} Our ablation study reveals important insights about the synergistic effects of our proposed components. While the attention mechanism alone provides a 1.1\% improvement in whole tumor Dice score, and prompting alone shows minimal improvement, the combination of both mechanisms yields a 2.3\% improvement. This suggests that prompting is most effective when guided by the refined features from the sub-region-aware attention mechanism, highlighting a synergistic relationship between the two components. The attention mechanism learns to emphasize informative modalities, while the prompting mechanism leverages this enhanced representation for iterative refinement.

\textbf{Sub-Region-Specific Performance.} The necrotic core remains the most challenging sub-region to segment, achieving a Dice score of 0.71 in our full framework compared to 0.64 in the baseline. While this represents a substantial 10.9\% improvement, the absolute performance reflects the inherent difficulty of this tissue type, which exhibits complex and heterogeneous appearance across modalities. In clinical practice, a Dice score of 0.71 for the necrotic core is clinically meaningful, as it significantly reduces manual correction effort and provides reliable guidance for treatment planning. The substantial improvement over baseline methods demonstrates the effectiveness of our sub-region-aware approach in addressing this challenging segmentation task.

\section{Discussion and Insights}

\textbf{Strengths.} Our work introduces a novel framework for adapting foundation models to multi-modal medical imaging, featuring sub-region-aware modality attention and adaptive prompt engineering. This framework provides a principled approach to tackling the challenges of multi-modal fusion and sub-region heterogeneity, moving beyond simple concatenation and fine-tuning. Our experimental results on the BraTS 2020 dataset validate the effectiveness of our approach, demonstrating significant improvements in segmentation accuracy, particularly for the challenging necrotic core sub-region. The sub-region-aware attention mechanism effectively learns to weight modalities based on their relevance to specific tumor components, providing interpretability into the model's decision-making process.

\textbf{Key Findings.} Our analysis yields several key insights. First, we confirm that naive multi-modal fusion is insufficient for complex medical imaging tasks, and that fine-tuning is a necessary but not sufficient step. Second, we demonstrate that a more sophisticated, attention-based fusion mechanism can unlock the full potential of multi-modal data. Finally, our work underscores the importance of sub-region-specific approaches, showing that a one-size-fits-all segmentation strategy is inadequate for heterogeneous tumors. These findings highlight the inherent complexity and variability in medical image segmentation, where different regions present distinct challenges and require adaptive processing strategies.

\textbf{Challenges.} While our proposed framework demonstrates significant promise, we acknowledge several limitations that open avenues for future research. First, our current approach processes 3D medical volumes on a slice-by-slice basis, which may not fully capture the rich 3D spatial relationships inherent in volumetric data. Second, our framework assumes the availability of all four MRI modalities. Developing strategies to handle missing or incomplete data would enhance the robustness and clinical applicability of the model. The absence of informative modalities introduces modality-related uncertainty \cite{alijani2025wqlcp}.
Finally, while we have validated our framework on the BraTS 2020 dataset, further evaluation on other datasets and across different institutions would be beneficial to assess its generalizability.

\textbf{Future Work.} Based on our findings, future research directions include: (1) developing true 3D processing strategies that leverage volumetric convolutions or other techniques to capture full 3D spatial context, (2) investigating adaptive loss functions that weight different sub-regions based on their segmentation difficulty, (3) exploring robust fusion mechanisms that gracefully handle missing or incomplete modalities, and (4) extending the framework to additional medical imaging tasks and datasets to assess generalizability and identify domain-specific challenges. Furthermore, the limitations identified in this work, such as the challenges of sub-region heterogeneity and slice-by-slice processing, suggest that future work could also explore methods for quantifying the reliability of segmentation predictions, which would provide clinicians with valuable information for decision-making.


\begin{figure}[h]
  \centering
  \includegraphics[width=\textwidth]{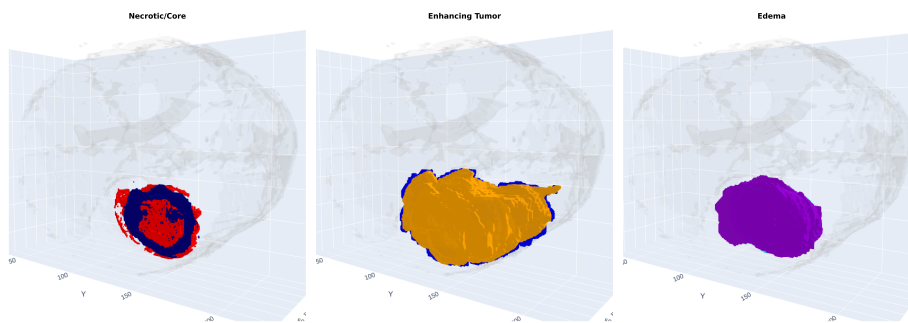} 
  \caption{Qualitative 3D visualization of sub-region-specific segmentation results from BraTS 2020. The subfigures show necrotic core (red), enhancing tumor (orange), and edema (purple) with ground truth (solid) and predictions (overlays).}
  \label{fig:results_viz}
\end{figure}

\section{Conclusion}

In this work, we introduced a novel framework for adapting foundation models to the complex task of multi-modal brain tumor segmentation. Our framework, which features sub-region-aware modality attention and adaptive prompt engineering, provides a principled and effective approach to multi-modal fusion and sub-region heterogeneity. Our experiments on the BraTS 2020 dataset demonstrate that our framework significantly outperforms baseline approaches, particularly in the segmentation of the challenging necrotic core. By moving beyond simple fine-tuning and introducing more sophisticated mechanisms for modality fusion and prompting, our work paves the way for the development of more accurate and robust foundation model-based solutions for a wide range of multi-modal medical imaging applications.

%
%

\end{document}